\pdfoutput=1
\documentclass[11pt]{article}

\usepackage[final]{acl}

\usepackage{times}
\usepackage{latexsym}
\usepackage{subfig}

\usepackage[T1]{fontenc}
\definecolor{cGrey}{HTML}{eff3fa} %#e6f9f6 %edf6f9 #ecf8ef #f0f6f9
\definecolor{cGreen}{HTML}{0077b6} %029e43
\newcommand{\pub}[1]{\color{gray}{\tiny{#1}}}
\newcommand{\model}{\textit{Sali4Vid}}
\newcommand{\videofeat}{\boldsymbol{x}}
\newcommand{\xspatial}{\videofeat^{spat}} % \boldsymbol{x}^{s}
\newcommand{\wvideofeat}{\hat{\videofeat}}

\newcommand{\segment}{\boldsymbol{S_m}}

\newcommand{\transcript}{\boldsymbol{y}}

\newcommand{\retrievedfeat}{\boldsymbol{r}}
\newcommand{\segmentretrievedfeat}{\tilde{\retrievedfeat}_{\segment}}

\usepackage{colortbl}
\usepackage{bm}
\usepackage{multirow}
\usepackage{amsmath}
\usepackage{pifont}
\usepackage{booktabs}
\usepackage{amssymb}
\usepackage{algorithm}
\usepackage{algorithmic}

\newcommand\blfootnote[1]{%
  \begingroup
  \renewcommand\thefootnote{}\footnote{#1}%
  \addtocounter{footnote}{-1}%
  \endgroup
}

\usepackage[utf8]{inputenc}

\usepackage{microtype}

\usepackage{inconsolata}

\usepackage{graphicx}

\title{Sali4Vid: Saliency-Aware Video Reweighting and \\  Adaptive Caption Retrieval for Dense Video Captioning}
\author{MinJu Jeon\quad
        Si-Woo Kim\quad
        Ye-Chan Kim\quad
        HyunGee Kim\quad
        Dong-Jin Kim$^{\dagger}$ \\
        Hanyang University, South Korea. \\
        {\footnotesize{\texttt{\{mnju5026, boreng0817, dpcksdl78, khjiiii2002, djdkim\}@hanyang.ac.kr}}}}

\AtEndPreamble{
    \usepackage[capitalize]{cleveref}
    \crefname{section}{Sec.}{Secs.}
    \Crefname{section}{Section}{Sections}
    \Crefname{table}{Table}{Tables}
    \crefname{table}{Tab.}{Tabs.}
    \Crefname{figure}{Figure}{Figures}
}

\begin{document}
\maketitle
\begin{abstract}
Dense video captioning aims to temporally localize events in video and generate captions for each event. While recent works propose end-to-end models, they suffer from two limitations: (1) applying timestamp supervision only to text while treating all video frames equally, and (2) retrieving captions from fixed-size video chunks, overlooking scene transitions. To address these, we propose $\model$, a simple yet effective saliency-aware framework. We introduce \textit{Saliency-aware Video Reweighting}, which converts timestamp annotations into sigmoid-based frame importance weights, and \textit{Semantic-based Adaptive Caption Retrieval}, which segments videos by frame similarity to capture scene transitions and improve caption retrieval. $\model$ achieves state-of-the-art results on YouCook2 and ViTT, demonstrating the benefit of jointly improving video weighting and retrieval for dense video captioning.\blfootnote{$^{\dagger}$Corresponding author.}\footnote{Code: \url{https://github.com/forminju/Sali4Vid}}
\end{abstract}

\section{Introduction}

\begin{figure}[t]
\begin{center}
\includegraphics[width=1\columnwidth]{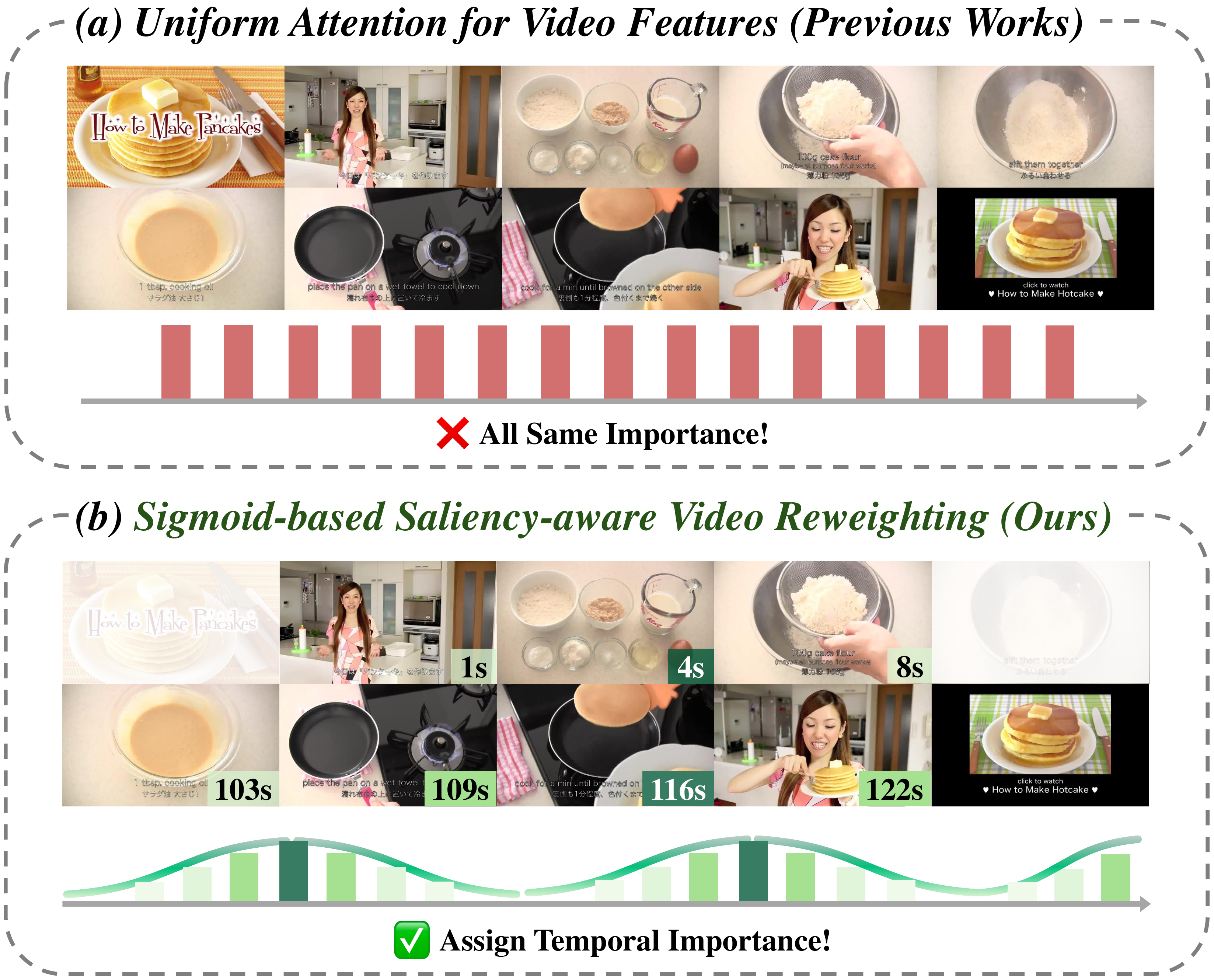}
\end{center}
\caption{(a) Previous works incorporate timestamps only on the textual side, treating all video features with uniform features. (b) We propose $\model$, a simple yet effective saliency-aware framework by leveraging sigmoid-based soft reweighting.}
\label{fig:overview}
\end{figure}

\begin{figure*}[t]
\begin{center}
\includegraphics[width=\linewidth]{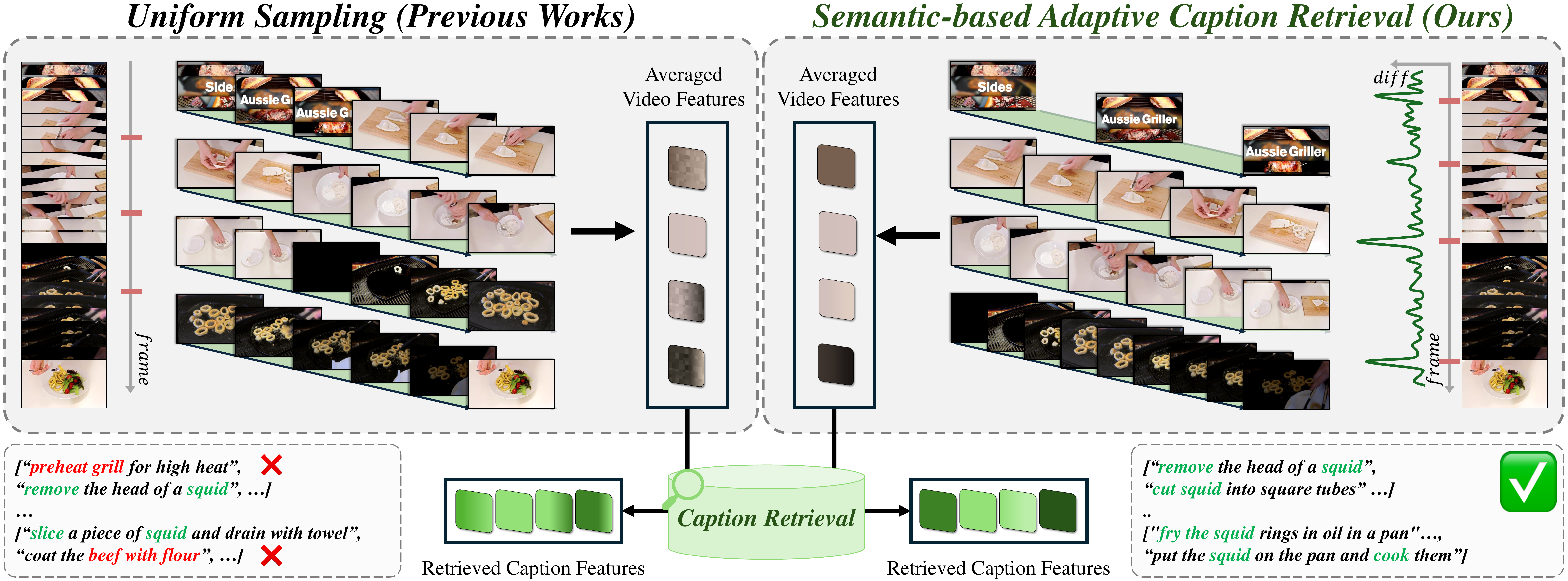}
\end{center}
\caption{(Left) The previous caption retrieval approach overlooks the scene transition, leading to redundant or misaligned captions that may not adequately reflect meaningful changes in the video content. (Right) $\model$ adaptively segments frames based on similarity difference, enabling more contextually aligned and diverse caption retrieval for meaningful segments.}
\label{fig:adaptive caption}
\end{figure*}

The dense video captioning (DVC) task~\cite{li2018dense, wei2023dense, duan2018dense, zhou2018dense, krishna2017dense, mkhallati2023dense} aims to localize multiple events in untrimmed videos and generate descriptive captions for each. Unlike standard video captioning (VC) ~\cite{gao2017video, chen2017video, wang2018video, seo2022video, zhao2023video, lee2024ifcap, kim2024semi}, which generates a single caption for a short and trimmed clip, DVC generates multiple temporally localized descriptions from long video streams, which is more challenging.
% dvc 설명

To effectively handle both localization and caption generation, prior works have proposed end-to-end modeling~\cite{pdvc, streaming}. For instance, Vid2Seq~\cite{vid2seq} formulates DVC as a sequence-to-sequence task, adding time tokens to the text for timestamp supervision. More recently, CM$^2$~\cite{cm2} and HiCM$^2$~\cite{hicm2} further extend this direction by retrieving auxiliary captions from an external datastore using video features as queries.
% 기존 방식 설명

Despite these advances, existing methods still suffer from two key limitations. \textbf{First}, although fully-supervised timestamp annotations are available in training, previous work leverages them only on the textual side, while treating the video features as uniformly important across time as shown in \Cref{fig:overview} (a). 
% This design does not fully leverage the available temporal annotations to guide video features. 
\textbf{Second}, recent caption retrieval methods~\cite{cm2, hicm2} adopt fixed-size clip-level retrieval for auxiliary captions. However, this strategy overlooks semantic transitions and scene changes within the video, which can lead to misaligned or redundant caption retrieval. For example, as illustrated in \Cref{fig:adaptive caption} (Left), unrelated actions like \textit{preheat grill} and \textit{remove squid} may be grouped into the same chunk, resulting in retrieved captions that may not accurately describe each event.

To address these limitations, we propose $\model$, a simple yet effective saliency-aware framework that explicitly applies temporal saliency cues to enhance video features during training and adaptively retrieves captions based on semantic transitions, providing more accurate event captions.

Specifically, we enhance video features by saliency reweighting based on timestamps, which emphasizes frames around annotated start and end points. 
This approach enables the model to directly utilize temporal supervision on the video side.
Consequently, the model effectively focuses on salient visual regions, as demonstrated in \Cref{fig:overview} (b). In addition, we calculate frame-to-frame similarity to find semantic transitions and segment the video adaptively, avoiding fixed-size clip-level retrieval that may group unrelated actions into the same chunk. This enables the retrieval of captions that are better aligned with each meaningful segment, as illustrated in \Cref{fig:adaptive caption} (Right).

Our model contains two main key components: \textit{Saliency-aware Video Reweighting} and \textit{Semantic-based Adaptive Caption Retrieval}. \textbf{First}, during the training phase, the \textit{Saliency-aware Video Reweighting} provides timestamp supervision to the visual side through sigmoid-based weights, allowing the model to continuously focus on salient video frames. \textbf{Second}, the \textit{Semantic-based Adaptive Caption Retrieval} segments the video based on frame-to-frame similarity differences, finding meaningful semantic changes, and retrieves captions aligned with these semantically adaptive segments.

Empirically, our framework achieves state-of-the-art results with a CIDEr score of 75.80 on YouCook2 and 53.32 on ViTT, outperforming the previous state-of-the-art by +3.96 on YouCook2 and +2.58 on ViTT.

We summarize our contributions as follows:
\begin{itemize}
    \setlength\itemsep{0em}
    \item We propose $\model$, a saliency-aware framework that enhances video features by applying sigmoid-based reweighting with timestamp supervision, focusing on more salient features in the training phase.
    \item We introduce a semantic-based adaptive caption retrieval strategy that segments videos based on frame-level similarity differences, enabling the retrieval of more contextually aligned captions for each semantic segment.
    \item We validate our method on YouCook2 and ViTT, achieving state-of-the-art results in both the localization and captioning tasks.
\end{itemize}

\section{Related Work}

\subsection{Dense Video Captioning}
%dvc 정의
Dense Video Captioning (DVC) aims to temporally localize events within untrimmed videos and generate captions for each event~\cite{dvc}. Early approaches typically have adopted a two-stage "localize-and-describe" pipeline~\cite{better, multi}. However, this separation between localization and captioning often overlooks the interaction between the two subtasks, leading to suboptimal performance.
To address this, recent works have explored end-to-end frameworks that jointly model event localization and captioning. PDVC~\cite{pdvc} reformulates DVC as a set prediction problem using a DETR-style transformer~\cite{carion2020end}, enabling parallel prediction of temporal segments and captions without relying on intermediate proposals.
More recently, Vid2Seq~\cite{vid2seq} formulates dense video captioning as a sequence-to-sequence task, generating both timestamp tokens and captions in a unified output while leveraging large-scale speech transcriptions. Building on this, Streaming V2S~\cite{streaming} introduces streaming decoding with visual memory for online captioning, and DIBS~\cite{dibs} proposes scalable pretraining with pseudo-labeled segments. CM$^2$~\cite{cm2} and HICM$^2$~\cite{hicm2} further extend this line of work by integrating retrieval-augmented generation using external caption memories.
Unlike previous methods that apply timestamp supervision only to text, overlooking video-side temporal modeling, our $\model$ explicitly leverages timestamp annotations to reweight video features and adaptively retrieves segment-level captions based on semantic transitions.

\begin{figure*}[t]
\centering\includegraphics[width=1\textwidth]{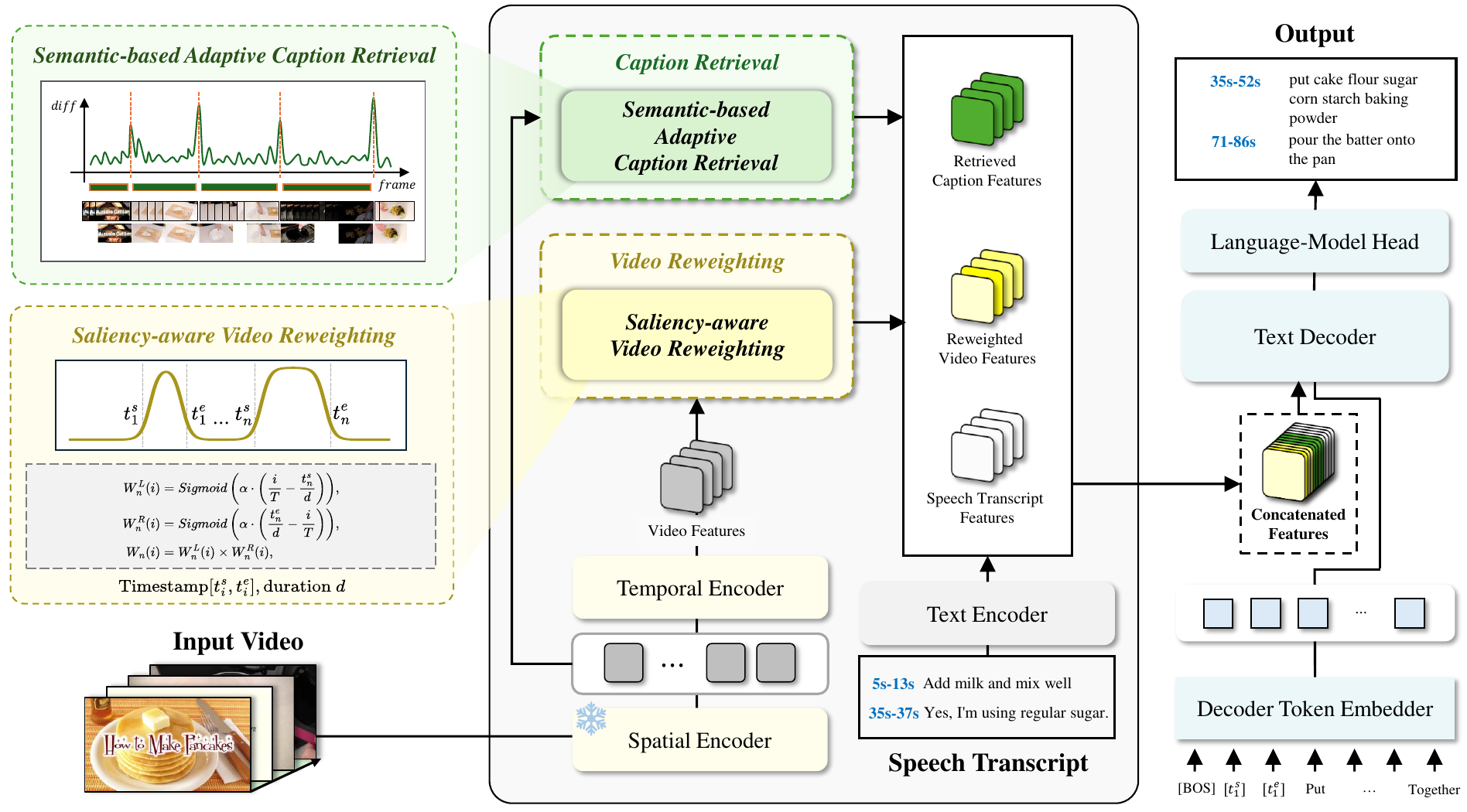}

\caption{Overview of our $\model$ framework for dense video captioning. $\model$ enhances dense video captioning by combining \textit{Saliency-Aware Video Reweighting} with \textit{Semantic-based Adaptive Caption Retrieval}. Specifically, we utilize timestamp supervision to softly reweight video features in the training phase and retrieve relevant captions by clustering semantically similar video frames. The reweighted video features $\wvideofeat$, segment-level retrieved caption features $\tilde{\retrievedfeat}$, and speech features $\transcript$ are then concatenated and passed through the cross-attention layer of the text decoder, enabling the model to better localize events and generate accurate event captions.}
\label{fig:main_figure}
\end{figure*}

\section{Proposed Method}
Recent work in dense video captioning~\cite{hicm2, dibs, vid2seq, streaming} often utilizes timestamp annotations only on the text side, while treating all video frames as equally important. To address these limitations, we propose $\model$, a framework that explicitly models frame-level importance through two complementary strategies, as illustrated in \Cref{fig:main_figure}.

\textit{First}, $\model$ applies sigmoid-based time stamp-guided weighting to highlight salient frames, providing explicit temporal supervision on the visual side during training.
\textit{Second}, our model captures semantic transitions by measuring frame-to-frame similarity, enabling adaptive segmentation for retrieving captions that better align with meaningful video segments. Together, these strategies improve event localization and caption generation by focusing on important visual content and retrieving contextually relevant captions. We detail these components in Section~\ref{sec:frame_selection} for saliency-aware video reweighting and Section~\ref{sec:caption_retrieval} for semantic-based adaptive caption retrieval.

\noindent\textbf{Preliminaries.}
We build on the structure of the Vid2Seq~\cite{vid2seq}, fine-tuning a model pre-trained on 1.8 million videos.
Given an input video, we extract frame-level features $\xspatial = \{\xspatial_i\}_{i=1}^{T}$ using CLIP ViT-L/14~\cite{radford2021learning,dosovitskiy2020image}.
Spatial features $\xspatial$ are processed by a temporal transformer to obtain context-aware video features $\boldsymbol{x} = \{\boldsymbol{x}_i\}_{i=1}^{T}$.

The goal is to predict a set of event segments and corresponding captions ${(t_n^s, t_n^e, C_n)}_{n=1}^{N}$, where $t_n^s$ and $t_n^e$ denote the start and end timestamps, and $C_n$ is the generated caption. Timestamps are normalized over the video duration $d$.

During training, we use ground-truth annotations $(t_n^s, t_n^e, C_n)$ and speech transcripts features $\transcript$. During inference, the model predicts event boundaries and captions without ground-truth timestamps, relying on video features $\boldsymbol{x}$ and transcript features $\transcript$, following the Vid2Seq setup.

\subsection{Saliency-aware Video Reweighting}
\label{sec:frame_selection}
Building on the observation that prior works~\cite{vid2seq, hicm2} treat video frames uniformly despite having timestamp annotations, we propose a \textit{Saliency-Aware Video Reweighting} method that directly leverages these annotations to compute frame-level importance scores as continuous, fully-supervised weights. Unlike methods that rely solely on textual cues~\cite{dibs, streaming} or require additional modules to infer saliency score~\cite{ge2025implicit}, our approach makes direct use of ground-truth event boundaries to provide continuous, fine-grained frame weighting in the training phase.

Specifically, we assign continuous sigmoid-based weights to each frame, as illustrated in \Cref{fig:main_figure}. For each annotated event $n$ with start and end times $(t_n^s, t_n^e)$, we define the sigmoid-based importance weight for frame $i$ as follows:

\begin{align}
{W_n^L(i)} &= Sigmoid\left( \alpha \cdot \left( \frac{i}{T} - \frac{t_n^s}{d} \right) \right), \\
{W_n^R(i)} &= Sigmoid\left( \alpha \cdot \left( \frac{t_n^e}{d} - \frac{i}{T} \right) \right), \\
{W_n(i)} &= {W_n^L(i)} \times{W_n^R(i)},
\end{align}
where $\alpha$ controls the sharpness of the sigmoid curve, and $\frac{t_n^s}{d}$, $\frac{t_n^e}{d}$ denote the normalized start and end of the $n$-th event.

If multiple events exist, the final importance weight is computed as:

\begin{equation}
W(i) = \max_n W_n(i).
\end{equation}

The frame feature $\boldsymbol{x}_i$ is then reweighted by the importance score as follows:

\begin{equation}
\wvideofeat_i = \boldsymbol{x}_i \cdot W(i),
\label{eq:masking}
\end{equation}
resulting in reweighted video features $\wvideofeat = \{\wvideofeat_i\}_{i=1}^{T}$ used for decoding.

Our sigmoid-based reweighting strategy enables the model to effectively highlight salient event regions by assigning continuous importance scores to each frame, capturing both central areas and temporal boundaries.
By providing smooth and fully supervised temporal guidance, it allows the model to focus on salient visual features, leading to improved caption generation and event localization, as illustrated in \Cref{fig:distribution}.

\subsection{Semantic-based Adaptive Caption Retrieval} 
\label{sec:caption_retrieval}
Recent studies~\cite{cm2, hicm2} suggest that auxiliary captions from an external datastore can provide useful semantic context for dense video captioning. However, these methods typically retrieve captions based on fixed-sized clip-level video features, overlooking the dynamic scene transitions within the video. When fixed-size clip-level video features are used as queries, multiple events may be mixed within a single clip. This makes it challenging to retrieve accurate captions that correctly match the target event. To address this limitation, we propose a \textit{Semantic-based Adaptive Caption Retrieval} that adaptively finds semantic segments and retrieves relevant captions for each segment.

\noindent{\textbf{Frame Difference Calculation.}}
We first compute frame-to-frame feature similarity differences to capture the semantic transitions. Given frame-level spatial features $\{\xspatial_i\}_{i=1}^{T}$, we calculate the cosine difference between consecutive frames as:
\begin{equation}
D(i) = 1 - \text{sim}\left( \xspatial_i, \xspatial_{i+1} \right),
\label{eq:frame-diff}
\end{equation}
where $\text{sim}(\cdot, \cdot)$ denotes cosine similarity. A large value of $D(i)$ signals a strong semantic change.
% $D(i)$ indicates the level of scene transition from frame $i$ and frame $i+1$. 
% 인접 프레임끼리 sim -> diff 계산

\noindent{\textbf{Adaptive Segment Construction.}} %Momentum-based 
After calculating the frame-level semantic difference $D(i)$, we segment the video based on $D(i)$. 
A straightforward method is to set a boundary whenever $D(i)$ exceeds a fixed threshold $\tau_{fixed}$.
However, this frame-wise segmentation is sensitive to small changes and noise, which can result in over-segmentation by splitting stable scenes into excessively short segments.

To address this issue, we adopt a \textit{momentum-based accumulation strategy} that continuously aggregates frame differences and captures boundaries only when a sustained change is observed. This approach reduces sensitivity to small variations and improves boundary detection by considering the accumulated difference shift across frames, as inspired by~\cite{kordopatis2019visil}.
% 
% These resulting segments are then used to retrieve semantically aligned auxiliary captions, providing fine-grained textual guidance for each localized video segment.

We first define the adaptive threshold $\tau_{adap}$ using the mean $\mu$ and standard deviation $\sigma$ of $\{D(i)\}_{i=1}^{T-1}$ as $\tau_{adap} = \mu + \beta \cdot \sigma$, where $\beta$ is a scaling factor.
% 인접 프레임 diff < tau : 누적, diff > tau : momentum update, starting index update
Starting from the first frame indexed by $s_{\text{cur}}=1$, we initialize the running segment feature with the first frame feature as $\boldsymbol{z}_\text{cur} = \xspatial_1$. We iteratively grow a segment by tracking $\boldsymbol{z}_{\text{cur}}$. For each subsequent frame $i+1$, we compute the semantic difference between the current segment feature and the incoming frame:
\begin{equation}
D'(i) = 1 - \boldsymbol{z}_{\text{cur}} \cdot \xspatial_{i+1},
\end{equation}
where $\xspatial_{i+1}$ is the spatial feature of frame $i+1$.

We apply the following decision rule:
If $D'(i) > \tau_{adap}$, the current segment is ended at $i$, a new segment starts at $i+1$, and the segment feature is reset to $\boldsymbol{z}_{\text{cur}} = \boldsymbol{x}_{i+1}^{spat}$.

Otherwise, we include frame index $i + 1$ in the current segment and update the segment feature by computing the moving average as:

\begin{equation}
\boldsymbol{z}_{\text{cur}} \leftarrow \frac{|\boldsymbol{S}_{\text{cur}}| \cdot \boldsymbol{z}_{\text{cur}} + \xspatial_{i+1}}{|\boldsymbol{S}_{\text{cur}}| + 1},
\
\end{equation}
where $|\boldsymbol{S}_{\text{cur}}|$ is the number of frames in the current segment.  % segment 에 대해서 평균 값 정하는 과정
This process continues until all frames are processed, yielding segments $\{\boldsymbol{S_1}, \boldsymbol{S_2}, \dots, \boldsymbol{S_m}\}$, where each segment is defined as the set of consecutive frame indices as $\segment = \{s_m, s_{m}+1, \dots, e_m\}$, with $s_m$ and $e_m$ denoting the start and end frame indices of the segment.
Each corresponding segment is computed as the average of the frame features within the segment as $\boldsymbol{z}_{\boldsymbol{S_m}} = \frac{\sum_{j=s_m}^{e_m} \boldsymbol{x}_j^{spat}}{|\boldsymbol{S_m}|}$, which serves representation for retrieving semantically aligned auxiliary captions.

\noindent{\textbf{Segment-level Caption Retrieval.}}
After segmenting the video into $\{\segment\}_{m=1}^{M}$, where $M$ is the total number of adaptively captured segments, we retrieve semantically aligned captions for each segment using its feature representation $\boldsymbol{z}_{\segment}$, providing localized textual guidance to each segment.

Given the segment-level feature $\boldsymbol{z}_{\segment}$, we compute similarity scores against the external caption datastore $R=\{\retrievedfeat_r\}_{r=1}^{N_R}$  and retrieve the $\operatorname*{Top-\textit{k}}$ semantically aligned captions as:
% $R = {\retrievedfeat_j}$ 
\begin{equation} 
\mathcal{R}_{\segment} = \operatorname*{Top-\textit{k}}_{\retrievedfeat_r \in R} \left( \text{sim}(\retrievedfeat_r, \boldsymbol{z}_{\boldsymbol{S_m}}) \right),
\label{eq:top-k}
\end{equation}
where $\text{sim}(\cdot, \cdot)$ denotes cosine similarity,
and $\mathcal{R}_{\segment} \in \mathbb{R}^{k \times D}$ represents the $\operatorname*{Top-\textit{k}}$ retrieved caption embeddings for $\segment$.
We then aggregate these retrieved caption embeddings by average pooling to obtain a single caption guidance vector:

\begin{equation}
\segmentretrievedfeat = \frac{1}{\textit{k}} \sum_{\retrievedfeat\in \mathcal{R}_{\segment}} \retrievedfeat,
\end{equation}
where $\segmentretrievedfeat \in \mathbb{R}^D$ is the averaged embedding representing the external semantic guidance for $\segment$.
We repeat this process for all segments, resulting in a set of segment-wise caption embeddings $\tilde{\retrievedfeat}=\{\segmentretrievedfeat\}_{m=1}^M$ that is used in decoding to provide textual guidance aligned with each segment.

\subsection{Model Training and Inference}
\label{sec:training}
Our model integrates reweighted video features $\wvideofeat$, segment-level retrieved caption features ${\tilde{\retrievedfeat}}$, and speech transcripts features $\transcript$ to improve event localization and caption generation. We extract frame features $\{\xspatial_i\}_{i=1}^{T}$, and perform segment-wise retrieval of $\operatorname*{Top-\textit{k}}$  caption embeddings ${\segmentretrievedfeat}$ from an external datastore for each segment.
The reweighted video features $\wvideofeat$ are obtained as described in \Cref{sec:frame_selection}.
Speech transcripts are encoded using a transformer-based text encoder with time tokens to obtain $\transcript$.

% We train the model using a cross-entropy loss conditioned on the reweighted video features $\wvideofeat$, retrieved captions ${\segmentretrievedfeat}$, and transcript features $\transcript$ to predict the target sequence $\boldsymbol{z}$:

We train the model using a cross-entropy loss conditioned on $\wvideofeat$, ${\tilde{\retrievedfeat}}$, $\transcript$ to predict the target sequence $\boldsymbol{o}$:

\begin{equation}
\mathcal{L}_{\theta} = \text{CE}(\boldsymbol{o} \mid \wvideofeat, \tilde{\retrievedfeat}, \transcript),
\end{equation}
where $\theta$ denotes model parameters. 
During inference, the model generates event-aware captions based on the video features, transcript features, and segment-level retrieved caption features. Unlike training, where annotated timestamps are used to apply importance weights, we perform inference \textit{without timestamp supervision and do not apply any weighting to the video features}.

\section{Experiment}
\begin{table*}[t]
\centering
% \scalebox{0.9}{
\resizebox{0.99\linewidth}{!}{
\begin{tabular}{l|c|cccc|cccc}
\toprule[2pt]
\multirow{2}{*}{\textbf{Method}} & \multirow{2}{*}{\textbf{PT}} & \multicolumn{4}{c|}{\textbf{YouCook2 (val)}} & \multicolumn{4}{c}{\textbf{ViTT (test)}} \\
\cmidrule(lr){3-6} \cmidrule(lr){7-10}
& & \textbf{CIDEr} & \textbf{METEOR} & \textbf{SODA\_c} & \textbf{BLEU4} & \textbf{CIDEr} & \textbf{METEOR} & \textbf{SODA\_c} & \textbf{BLEU4} \\
\midrule
PDVC~\pub{ICCV21} & \ding{55} & 29.69 & 5.56 & 4.92 & 1.40 & - & - & - & - \\
CM$^2$~\pub{CVPR24} & \ding{55} & 31.66 & 6.08 & 5.34 & 1.63 & - & - & - & - \\
\midrule
Streaming V2S~\pub{CVPR24} & \ding{51} & 32.90 & 7.10 & 6.00 & - & 25.20 & 5.80 & 10.00 & - \\
DIBS~\pub{CVPR24} & \ding{51} & 44.44 & 7.51 & 6.39 & - & - & - & - & - \\
Vid2Seq$^\dagger$~\pub{CVPR23} & \ding{51} & 66.29 & 12.41 & 9.87 & 5.64 & 48.84 & 9.51 & 14.99 & 0.71 \\
HiCM$^2$~\pub{AAAI25} & \ding{51} & \underline{71.84} & \underline{12.80} & \textbf{10.73} & \underline{6.11} & \underline{51.29} & \underline{9.66} & \underline{15.07} & \underline{0.86} \\
\rowcolor{cGrey}\textbf{Ours} & \ding{51} & \textbf{75.80} & \textbf{13.54} & \underline{10.28} & \textbf{6.35} & \textbf{53.87} & \textbf{10.05} & \textbf{15.08} & \textbf{0.91} \\
\bottomrule[2pt]
\end{tabular}}
\caption{Comparison with state-of-the-art methods on YouCook2 validation and ViTT test sets. PT indicates whether the model is pretrained. Bold and underline denote the best and second-best scores, respectively. "-" indicates unavailable results. $\dagger$ denotes results reproduced from official implementations. Our method achieves state-of-the-art performance in most of the metrics.}
\label{tab:sota-dense-captioning}
\end{table*}

\subsection{Experimental Settings}
\label{subsec: setting}
\noindent\textbf{Datasets.} \textbf{YouCook2}~\cite{youcook2} consists of 2,000 untrimmed videos. On average, 320 seconds and 7.7 localized sentences per video. \textbf{ViTT}~\cite{vitt} includes 8,000 untrimmed instructional videos averaging 250 seconds and annotated with 7.1 localized short tags.

\begin{table}[t]
    \centering
    \resizebox{\linewidth}{!}{ % 혹은 \textwidth
        \begin{tabular}{l|c|ccc|ccc}
        \toprule[2pt]
        \multirow{2}{*}{\textbf{Method}} & \multirow{2}{*}{\textbf{PT}} & \multicolumn{3}{c|}{\textbf{YouCook2 (val)}} & \multicolumn{3}{c}{\textbf{ViTT (test)}} \\
        \cmidrule(lr){3-5} \cmidrule(lr){6-8}
        & & \textbf{F1} & \textbf{Recall} & \textbf{Precision} & \textbf{F1} & \textbf{Recall} & \textbf{Precision} \\
        \midrule
        PDVC & \ding{55} & 26.81 & 22.89 & 32.37 & - & - & - \\
        CM$^2$ & \ding{55} & 28.43 & 24.76 & 33.38 & - & - & - \\
        \midrule
        Streaming V2S & \ding{51} & 24.10 & - & - & 35.40 & - & -  \\
        DIBS & \ding{51} & 31.43 & 26.24 & \textbf{39.81} & - & - & -  \\
        Vid2Seq$^\dagger$ & \ding{51} & 31.08 & 30.38 & 31.81 & \underline{46.21} & \textbf{45.89} & 46.53 \\
        HiCM$^2$ & \ding{51} & \underline{32.51} & \textbf{32.51} & 32.51 & 45.98 & 45.00 & \underline{47.00} \\
        \rowcolor{cGrey}
        \textbf{Ours} & \ding{51} & \textbf{33.61} & \underline{31.11} & \underline{36.57} & \textbf{46.58} & 44.31 & \textbf{49.10} \\
        \bottomrule[2pt]
    \end{tabular}}
    \caption{
    Localization results on YouCook2 validation and ViTT test sets. Bold denotes the best performance, and underline denotes the second-best performance. “-” result is unavailable.}
    \label{tab:localization}
\end{table}

\noindent\textbf{Evaluation Metrics.} 
We evaluate our method on two sub-tasks in DVC. By using the official evaluation tool~\cite{wang2020dense}, we use CIDEr~\cite{vedantam2015cider}, BLEU4~\cite{papineni2002bleu}, and METEOR~\cite{banerjee2005meteor} metrics, which compare the generated captions to the ground truth across IoU thresholds of (0.3, 0.5, 0.7, 0.9). Additionally, to assess storytelling ability, we use the SODA\_c metric~\cite{fujita2020soda}. For event localization, we calculate the average precision, average recall, and F1 score, averaging these metrics over IoU thresholds of (0.3, 0.5, 0.7, 0.9).

\noindent\textbf{Implementation Details.} 
Following previous works~\cite{vid2seq, hicm2}, we build upon the Vid2Seq model that is pre-trained on 1.8M videos, which uses the T5-Base model~\cite{raffel2020exploring}  as both the text encoder and decoder. Video frames are extracted at 1 FPS and sub-sampled or padded to a fixed length $T=100$. The model is first trained for 10 epochs following Vid2Seq with a learning rate of 3e-4, then fine-tuned for 10 more epochs with our method using a learning rate of 1e-6, linearly warmed up over the first 10\% of steps and decayed to 0 via a cosine schedule. Training is performed on a single A6000 GPU with batch size 8, taking approximately 1h 20m total (4m 20s/epoch). We set $\alpha=10.0$ for sigmoid reweighting and $\beta=1.0$ for adaptive segmentation. Hyperparameter details are provided in the supplementary. The caption retrieval datastore is constructed from the training captions only.

\begin{table*}[t]
  \centering
  \resizebox{0.99\linewidth}{!}{
    % \begin{tabular*}{0.99\textwidth}{@{\extracolsep{\fill}}lccccccc@{}}
    \begin{tabular}{l|cccc|ccc}
    \toprule[2pt]
    \multirow{2}{*}{\textbf{Method}} & \multicolumn{4}{c|}{\textbf{Captioning}} & \multicolumn{3}{c}{\textbf{Localization}} \\
    \cmidrule(lr){2-5} \cmidrule(lr){6-8}
      & \textbf{CIDEr} & \textbf{METEOR} & \textbf{SODA\_c} & \textbf{BLEU4}
      & \textbf{F1} & \textbf{Recall} & \textbf{Precision} \\
    \midrule
    Baseline & 66.29 & 12.41 & 9.87 & 5.64 & 31.08 & 30.38 & 31.81 \\
    \midrule

    \rowcolor{cGrey}
    \multicolumn{8}{l}{\emph{Different Design of Saliency-Weights}} \\
    Hard binary mask & 68.85 & 12.53 & 10.25 & 5.95 & 32.53 & \textbf{33.06} & 32.03 \\
    Gaussian weights & 68.66 & 12.49 & 10.30 & 5.93 & 32.25 & 32.81 & 31.72  \\
    Sigmoid weights & \textbf{74.72} & \textbf{13.43} & \textbf{10.35} & \textbf{6.01} & \textbf{33.34} & 31.24 & \textbf{35.76} \\
    \midrule
    \rowcolor{cGrey}
    \multicolumn{8}{l}{\emph{Different Segment Feature Design for Caption Retrieval}} \\
    % \multicolumn{8}{l}{\textbf{\emph{Semantic Change Based Caption Retrieval}}} \\
     Mean-Pool & \textbf{75.80} & \textbf{13.54} & \textbf{10.28} & \textbf{6.35} & \textbf{33.61} & \textbf{31.11} & \textbf{36.57} \\
    Max-Pool & 75.02 & 13.48 & 10.22 & 6.23 & 33.01 & 30.37 & 36.17\\
    Key-Frame & 74.87 & 13.41 & 10.16 & 6.26 & 33.12 & 30.42 & 36.34 \\
    \bottomrule[2pt]
    \end{tabular}
    }
  \caption{Component-wise results on the YouCook2 validation set for both captioning and localization tasks. Sigmoid-based weights achieve the best performance among reweighting strategies, while mean-pooling caption retrieval shows the best results among the retrieval designs.}
  \label{tab:ablation-mask}
\end{table*}
\begin{table}[t]
    \centering
    \resizebox{0.45\textwidth}{!}{
        \begin{tabular}{l|cccc}
            \toprule[2pt]
            \multirow{2}{*}{\textbf{Mask Design}}  & \multicolumn{4}{c}{\textbf{YouCook2}} \\
             & \textbf{C} & \textbf{M} & \textbf{S\_c} & \textbf{F1} \\
            \midrule
            Baseline & 66.29 & 12.41 & 9.87 & 31.08 \\
        \midrule
            Start skew & 67.45 & 12.44 & 10.29 & 32.22  \\
            End skew & 67.54 & 12.47 & \textbf{10.31} & 32.22 \\
            Random skew & 67.06 & 12.27 & 10.27 & 32.61 \\
            \rowcolor{cGrey} Center skew (Ours)  & \textbf{75.80} & \textbf{13.54} & 10.28 & \textbf{33.61}\\
            \bottomrule[2pt]
        \end{tabular}
    }
    \caption{Ablation study of various mask designs in sigmoid-based importance modeling with adaptive caption retrieval.}
    \label{tab:rebut_design_sigmoid}
\end{table}
\begin{table}[t]
    \centering
    \resizebox{0.45\textwidth}{!}{
        \begin{tabular}{l|cccc}
            \toprule[2pt]
            \multirow{2}{*}{\textbf{Retrieval Design}}  & \multicolumn{4}{c}{\textbf{YouCook2}} \\
             & \textbf{C} & \textbf{M} & \textbf{S\_c} & \textbf{F1} \\
            \midrule
            Baseline & 66.29 & 12.41 & 9.87 & 31.08 \\
            % Fixed-size~\citeyearpar{hicm2} & 67.90 & 12.49 & 10.38 & 32.31  \\
        Fixed-size Clip-level & 63.96  & 12.14 & 9.93 & 32.24  \\
\midrule
            $\tau_{fixed}$ + \texttt{w/o MMT} & 66.05 & 12.33 & 10.23 & 32.73 \\
            $\tau_{fixed}$ + \texttt{MMT} & 66.87 & 12.26 & 10.32 & 32.49 \\
            $\tau_{adap}$ +  \texttt{w/o MMT} & 66.65 & 12.44 & 10.20 & 32.76 \\
            \rowcolor{cGrey}$\tau_{adap}$ + \texttt{MMT} (Ours)  & \textbf{68.63} & \textbf{12.61} & \textbf{10.33} & \textbf{32.79}\\
            \bottomrule[2pt]
        \end{tabular}
    }
    \caption{Ablation study on different designs for caption retrieval \textit{without} video reweighting. \texttt{MMT} denotes the momentum-based accumulation strategy. For the fixed-size setting, we set the window size to 10. C, M, and S\_c denote CIDEr, METEOR, and SODA\_c, respectively.}
    \label{tab:retrieval-design}
\end{table}

\begin{figure}[t]
\centering\includegraphics[width=0.99\columnwidth]{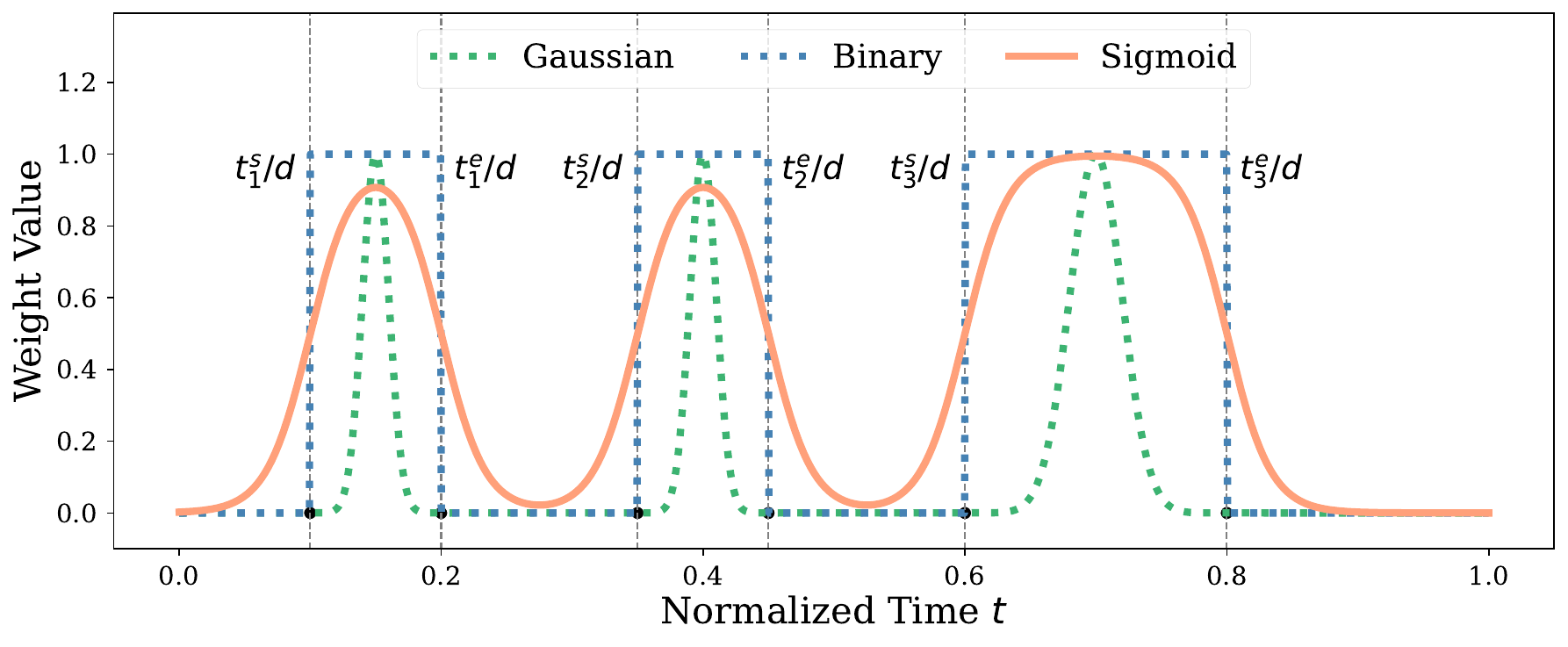}
\caption{Comparison of different weights with multiple-timestamps. Unlike Gaussian or binary, our sigmoid-based weight provides continuous importance weights while preserving the start and end boundaries.}
\label{fig:distribution}
\end{figure}

\noindent\textbf{Comparison with State-of-the-Arts.}
\Cref{tab:sota-dense-captioning} and \Cref{tab:localization} summarize the results on YouCook2 and ViTT. Our $\model$ achieves the best overall performance in both captioning and localization tasks. Specifically, in \Cref{tab:sota-dense-captioning}, our method achieves a CIDEr of \textbf{75.80} on YouCook2 and \textbf{53.87} on ViTT, outperforming prior state-of-the-art methods such as HiCM$^2$~\cite{hicm2} and Vid2Seq~\cite{vid2seq} across most metrics. This improvement stems largely from our use of sigmoid-based video reweighting and semantic-based adaptive caption retrieval, which allows the model to generate more accurate and semantically aligned captions for each localized region.

In \Cref{tab:localization}, $\model$ achieves the highest F1 scores of \textbf{33.61} on YouCook2 and \textbf{46.58} on ViTT. We also observe an improvement in precision over the baseline, +4.76 on YouCook2 and +2.57 on ViTT, demonstrating the effectiveness of applying supervision directly to video features during training via our saliency-aware reweighting strategy.

\subsection{Ablation Study}
We conduct ablation studies on the YouCook2 validation set to analyze the contributions of each proposed component, including saliency-aware video reweighting, semantic-based adaptive caption retrieval, and their combined impact on performance.

\noindent\textbf{Saliency-aware Video Reweighting.}
In \Cref{tab:ablation-mask}, we compare different importance weighting strategies. Sigmoid-based weighting achieves the best performance with a 74.72 CIDEr, significantly outperforming binary (68.85) and Gaussian (68.66) approaches and yielding a +8.43 CIDEr improvement over the baseline. We hypothesize that the sigmoid's smooth transitions around annotated event boundaries are particularly effective in our fully supervised setting. This property also substantially improves precision to 35.76 (vs. 32.03 for binary) by reducing false positives from overly sharp boundary decisions, as shown in \Cref{fig:distribution}.

We further investigate sigmoid mask designs in \Cref{tab:rebut_design_sigmoid}, comparing our proposed center-skew design against variants that emphasize the start, end, or random regions of an event. The center-skew mask consistently outperformed others, demonstrating that emphasizing the central region while preserving boundary information is most aligned with the structure of instructional videos.

% \noindent\textbf{Various Designs in Sigmoid-based Mask.}

\begin{table}[t]
    \centering
    \resizebox{0.45\textwidth}{!}{
        \begin{tabular}{c|cccc}
            \toprule[2pt]
            \textbf{$k$ Retrieved} & \multicolumn{4}{c}{\textbf{YouCook2}} \\
            \textbf{Captions} & \textbf{C} & \textbf{M} & \textbf{S\_c} & \textbf{F1} \\
            \midrule
            5 & 75.22 & 13.50 & \textbf{10.39} & 33.51 \\
            \rowcolor{cGrey}\textbf{10} & \textbf{75.80} & \textbf{13.54} & 10.28 & 33.61 \\
            20 & 75.51 & 13.52 & 10.26 & 33.58 \\
            30 & 75.52 & 13.52 & 10.29 & \textbf{33.65}  \\
            \bottomrule[2pt]
        \end{tabular}
    }
    \caption{Ablation study on different numbers of captions used for caption retrieval.}
    \label{tab:retrieval-num}
\end{table}

\begin{table}[t]
    \centering
    \resizebox{0.45\textwidth}{!}{
        \begin{tabular}{l|cccc}
            \toprule[2pt]
            \multirow{2}{*}{\textbf{Data Stores}} & \multicolumn{4}{c}{\textbf{YouCook2}} \\
             & \textbf{C} & \textbf{M} & \textbf{S\_c} & \textbf{F1} \\
            \midrule
            COCO~\citeyearpar{lin2014microsoft} & 75.51 & \textbf{13.54} & 10.44 & 33.59 \\
            CC3M~\citeyearpar{changpinyo2021conceptual} & 75.33 & 13.51 & 10.38 & 33.53 \\
            Hierarchical~\citeyearpar{hicm2} & \textbf{76.77} & 13.38 & \textbf{10.57} & 32.92 \\
            \rowcolor{cGrey}\textbf{In-domain} & 75.80 & \textbf{13.54} & 10.28 & \textbf{33.61} \\
            \bottomrule[2pt]
        \end{tabular}
    }
    \caption{Ablation study on different datastores used for caption retrieval.}
    \label{tab:retrieval-datastore}
\end{table}

\begin{table}[t]
    \centering
    \resizebox{0.45\textwidth}{!}{
        \begin{tabular}{cc|cccc}
            \toprule[2pt]
            \multirow{2}{*}{\textbf{Reweight}} & \multirow{2}{*}{\textbf{Cap}} & \multicolumn{4}{c}{\textbf{YouCook2}} \\
             & & \textbf{C} & \textbf{M} & \textbf{S\_c} & \textbf{F1} \\
            \midrule
            \ding{55} & \ding{55} & 66.29 & 12.41 & 9.87  & 31.08 \\
            \ding{55} & \ding{51} & 68.63 & 12.61 & 10.33  & 32.79   \\
            \ding{51} & \ding{55} & 74.72 & 12.49 & \textbf{10.35} & 33.34 \\
            \rowcolor{cGrey}\ding{51} & \ding{51} & \textbf{75.80} & \textbf{13.54} & 10.28 & \textbf{33.61} \\
            \bottomrule[2pt]
        \end{tabular}
    }
    \caption{Ablation study on our key components on YouCook2. \textbf{Reweight} denotes saliency-aware video reweighting, and \textbf{Cap} denotes semantic-based adaptive caption retrieval. Applying both components achieves the best performance.}
    \label{tab:component}
\end{table}

\begin{table}[t]
\centering
    \resizebox{0.45\textwidth}{!}{%
\begin{tabular}{lccc}
\toprule[2pt]
Method & Positive ($\uparrow$) & Negative ($\downarrow$) & IoU@0.1 ($\uparrow$) \\
\midrule
Baseline~\citeyearpar{vid2seq} & 0.23 & 0.21 & 0.038 \\
\rowcolor{cGrey} \textbf{Ours}& \textbf{0.25} & \textbf{0.20} & \textbf{0.049} \\
\bottomrule[2pt]
\end{tabular}
}
\caption{Attention score and IoU comparison during inference. Positive and Negative denote the average attention within and outside ground-truth segments, while IoU@0.1 measures overlap between the top-10\% attention regions and ground truth. }
\label{tab:attn_iou}
\end{table}
\noindent\textbf{Semantic-based Adaptive Caption Retrieval.}
\Cref{tab:ablation-mask} shows that mean-pooling performs best among aggregation strategies, improving CIDEr to \textbf{75.80} and F1 to \textbf{33.61}.
\Cref{tab:retrieval-design} further demonstrates the effectiveness of our momentum-based accumulation combined with adaptive thresholding without video reweighting, achieving the highest performance with a CIDEr of \textbf{68.63} and an F1 of \textbf{32.79}.
In contrast, the fixed-size setting shows slightly lower performance and needs extensive window tuning across datasets. Our semantic-based adaptive retrieval alleviates this by using an adaptive thresholding strategy with a single scaling factor~$\beta$.

We also analyze the number of retrieved features in \Cref{tab:retrieval-num}, where retrieving 10 captions per segment performs best.
Additionally, \Cref{tab:retrieval-datastore} shows that our approach is robust across datastores, while in-domain captions yield the highest F1 score, out-of-domain datastores like COCO~\cite{lin2014microsoft} and CC3M~\cite{changpinyo2021conceptual} produce comparable results, and hierarchical memory~\cite{hicm2} boosts both CIDEr and SODA\_c.

\noindent\textbf{Component Ablation.}
% table 7
In \Cref{tab:component}, we analyze the contribution of each component. We observe that applying adaptive caption retrieval alone improves CIDEr by \textbf{+2.34} and F1 by \textbf{+1.71}, showing that adaptive caption retrieval is beneficial to both captioning and localization. Saliency-aware reweighting alone improves CIDEr by \textbf{+8.43} and F1 by \textbf{+2.26} compared to baseline, confirming the importance of focusing on informative frames. Combining both achieves the best performance, improving all metrics. These results show that the two components are complementary and most effective when applied together.

\begin{figure}[t]
\centering\includegraphics[width=0.99\columnwidth]{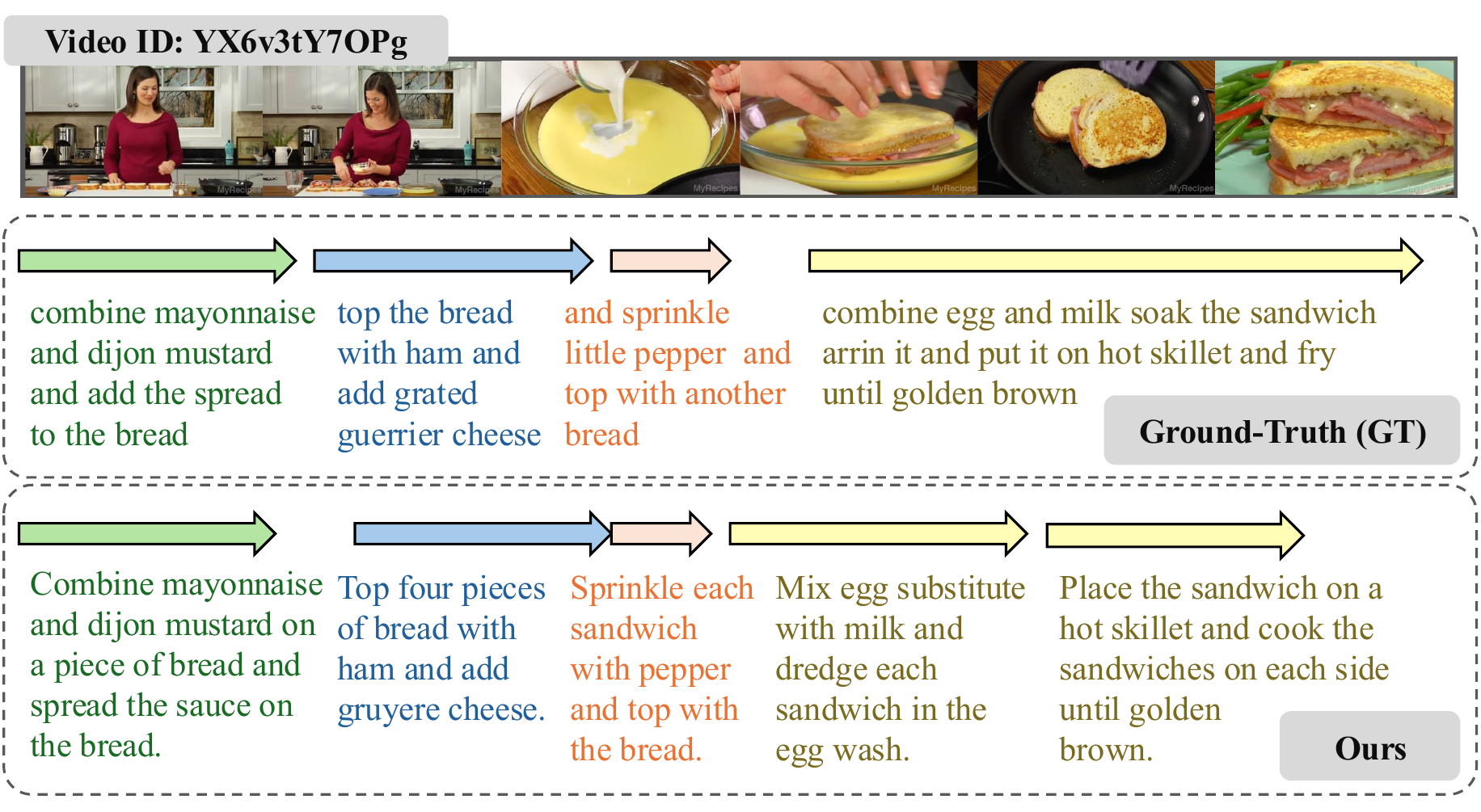}
\caption{A qualitative result from YouCook2 validation set.}
\label{fig:qualitative}
\end{figure}

\begin{table}[t]
    \centering
    \resizebox{0.45\textwidth}{!}{%
        \begin{tabular}{c|c|cc|ccc}
            \toprule[2pt]
            \textbf{Subset} & \textbf{\# Caps} & \textbf{Segment}& \textbf{Search}& \textbf{C} & \textbf{S\_c} & \textbf{F1} \\
            \midrule
            0\%   & 0 & - & - & 74.72 & \textbf{10.35} & 33.34 \\
            10\%   & 0.96K& 16.02 & 2.07  & 74.92 & 10.17 & 33.27  \\
            30\%  & 2.8K  & 16.02 & 2.09 & 75.67 & 10.25  &\textbf{33.62} \\
            100\%  & 9.6K & 16.02 & 2.11 &\textbf{75.80} & 10.28 & 33.61 \\
            \rowcolor{cGrey}
            \bottomrule[2pt]
        \end{tabular}
    }
    \caption{
        Efficiency-performance trade-off under varying relevant caption subset sizes (ms/vid). 
    }
    \label{tab:efficiency}
\end{table}

\noindent\textbf{Efficiency of Caption Retrieval.}
\Cref{tab:efficiency} presents the efficiency-performance trade-off under different retrieval subset sizes. Notably, even a small subset achieves comparable performance to the full set, while reducing retrieval cost to around 2 ms per video. In addition, segmentation requires about 16 ms per video, which is not negligible but performed only once using a lightweight, model-free clustering algorithm based on frame similarity. We will revise our manuscript to include this segmentation time report, further demonstrating the overall efficiency of our retrieval strategy.

\noindent\textbf{Analysis of Train-Test Mismatch Setting.}
During training, our saliency-aware video reweighting leverages timestamp supervision to guide the model, but not at test time. We regard this process as representation learning: with stronger supervision during training, the model learns more informative features and thus performs better at inference without extra components. To validate this, we compare attention maps from the last layer of the temporal encoder between Baseline~\cite{vid2seq} and our method in ~\Cref{tab:attn_iou}. The results show that our training strategy improves temporal focus, supporting accurate captioning during inference.
% The goal of our saliency-aware video reweighting module is to guide the model to learn more informative representations that enable precise localization and captioning during inference. To support this, we visualize the attention maps from the visual temporal encoder layer during inference as shown in \Cref{tab:attn_iou}. Positive and Negative in the table denote the average attention scores within and outside ground-truth timestamps. IoU@0.1 measures the overlap between the top-10\% attention regions and ground-truth segments. While this module is only applied during training, it does not cause train-test mismatch, as its role is to enhance representation learning rather than affect inference-time behavior directly.

\noindent\textbf{Qualitative Results.}
\Cref{fig:qualitative} shows examples from the YouCook2 validation set. Our $\model$ predicts event boundaries and captions that align well with the video content. For instance, it separates two closely occurring events (67–98s; yellow arrow) into distinct segments, illustrating its ability to capture detailed event transitions.

\begin{figure}[t]
\centering\includegraphics[width=1.0\columnwidth]{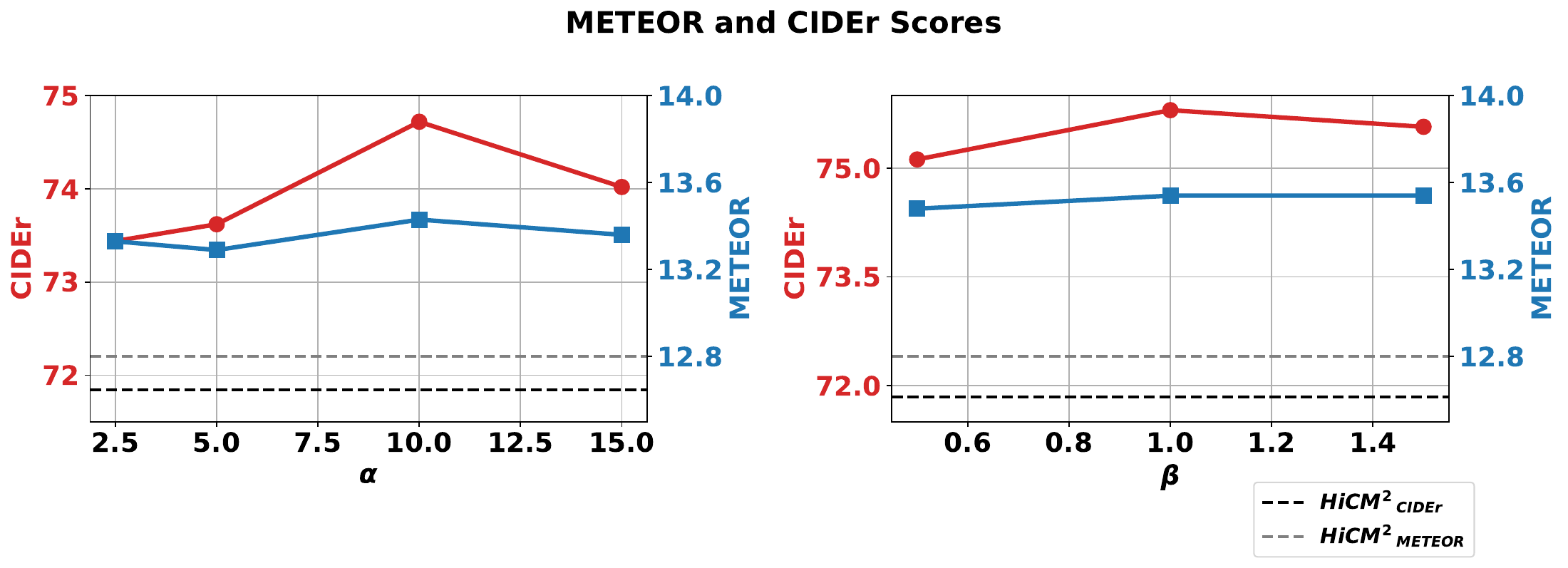}
\caption{Impact of hyper-parameter $\alpha$ for video reweighting and $\beta$ for semantic-based caption retrieval on captioning performance.}
\vspace{-1mm}
\label{fig:hyperparam_cider}
\end{figure}

\begin{figure}[t]
\centering\includegraphics[width=1.0\columnwidth]{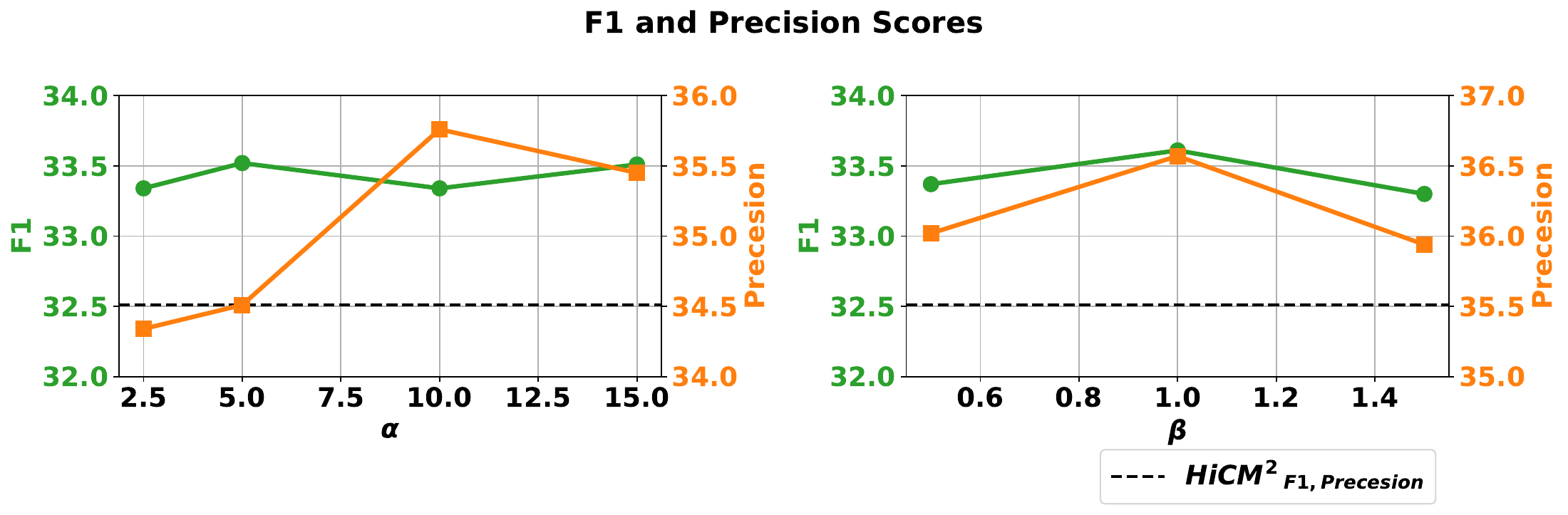}
\caption{Impact of hyper-parameter $\alpha$ for video reweighting and $\beta$ for semantic-based caption retrieval on localization performance.}
\vspace{-1mm}
\label{fig:hyperparam_localize}
\end{figure}
\label{supp:hyperparameters}

% \noindent\textbf{Hyperparameter Choice.}
% We have conducted ablation studies to evaluate the impact of two key parameters: the sharpness-controlling factor $\alpha$ for the sigmoid-based curve, and the scaling factor $\beta$, which determines $\tau_{adap}$. 
% First, \Cref{fig:hyperparam_cider} illustrates the variation in CIDEr and METEOR scores with respect to changes in the hyper-parameters $\alpha$ and $\beta$.
% The results demonstrate that, regardless of the hyperparameter settings, all CIDEr and METEOR scores consistently exceed those of the existing state-of-the-art method~\cite{hicm2}.
% Second, \Cref{fig:hyperparam_localize} shows how variations in the hyper-parameters $\alpha$ and $\beta$ influence the F1 and Precision scores in localization performance.
% Similarly with previous results, regardless of the hyper-parameters setting, most localization scores consistently surpass~\cite{hicm2}.
% We employ the hyperparameter setting of $\alpha=10$, $\beta=1.0$, which yields the best overall performance.

\noindent\textbf{Hyperparameter Choice.}
We conducted ablation studies on two key parameters: the sharpness factor $\alpha$ for the sigmoid curve and the scaling factor $\beta$ that controls $\tau_{adap}$. \Cref{fig:hyperparam_cider} presents CIDEr and METEOR scores across different values of $\alpha$ and $\beta$, while \Cref{fig:hyperparam_localize} reports the corresponding F1 and Precision for localization. In all cases, our method maintains performance above the previous state-of-the-art~\cite{hicm2}, demonstrating robustness to hyperparameter variation. We adopt $\alpha=10$ and $\beta=1.0$, which yield the best results.

\section{Conclusion}
We propose $\model$, a framework that incorporates saliency-aware modeling via two complementary strategies. It consists of two key components: (1) \textit{saliency-aware video reweighting}, which leverages timestamp annotations to compute continuous frame-level saliency weights, and (2) \textit{semantic-based adaptive caption retrieval}, which captures meaningful scene transitions and retrieves more accurate captions aligned with these segments, suppressing irrelevant information. Extensive experiments on YouCook2 and ViTT demonstrate that $\model$ achieves state-of-the-art performance in both captioning and event localization. Moreover, this framework can be readily extended to a wide range of vision-language modeling tasks~\cite{preserving, sync, verbdiff, sida, vipcap} beyond video captioning.
% \section{Limitation}
% \label{sec:limitation}
% Our model, $\model$, achieves state-of-the-art performance on dense video captioning by leveraging annotation-based video reweighting and semantic difference-based adaptive caption retrieval. However, limitations remain. The reweighting strategy depends on timestamp annotation, limiting applicability to weakly supervised or annotation-free settings. Moreover, as shown in \Cref{fig:semantic-analysis} in the supplementary, the adaptive retrieval module can still produce noisy segments. In future work, we plan to explore supervision-free reweighting and improve the robustness of adaptive retrieval.
\section{Limitation} 
\label{sec:limitation} 
Our model, $\model$, achieves state-of-the-art performance on dense video captioning through annotation-based video reweighting and semantic difference-based adaptive caption retrieval. However, some limitations remain. The reweighting strategy relies on timestamp annotation, reducing applicability to weakly supervised or annotation-free settings. Moreover, as shown in \Cref{fig:semantic-analysis} of the supplementary, the adaptive retrieval module may still yield noisy segments. In future work, we plan to explore supervision-free reweighting and enhance robustness of adaptive retrieval.

\section{Acknowledge}
\label{sec:acknowledge}
  This was partly supported by the Institute of Information \& Communications Technology Planning \& Evaluation (IITP) grant funded by the Korean government(MSIT) (No.RS-2020-II201373, Artificial Intelligence Graduate School Program(Hanyang University)) and the Institute of Information \& Communications Technology Planning \& Evaluation (IITP) grant funded by the Korean government(MSIT) RS-2025-25422680, Metacognitive AGI Framework and its Applications).

\clearpage

\bibliography{custom}

\appendix

\clearpage
\section*{Appendix}
\label{sec:appendix}

\setcounter{page}{1}
\setcounter{table}{0}
\setcounter{figure}{0}
\renewcommand{\thetable}{A.\arabic{table}}
\renewcommand{\thefigure}{A.\arabic{figure}}

In this Appendix, we provide additional details and qualitative results to support our findings. Specifically, \S\ref{supp:semantic-analysis} offers an analysis of the semantic-based caption retrieval, and \S\ref{supp:qualitative} showcases further qualitative results of our model.

\section{More Analysis for Semantic-based Caption Retrieval}
\label{supp:semantic-analysis}

\begin{figure*}[h]
\centering\includegraphics[width=1\textwidth]{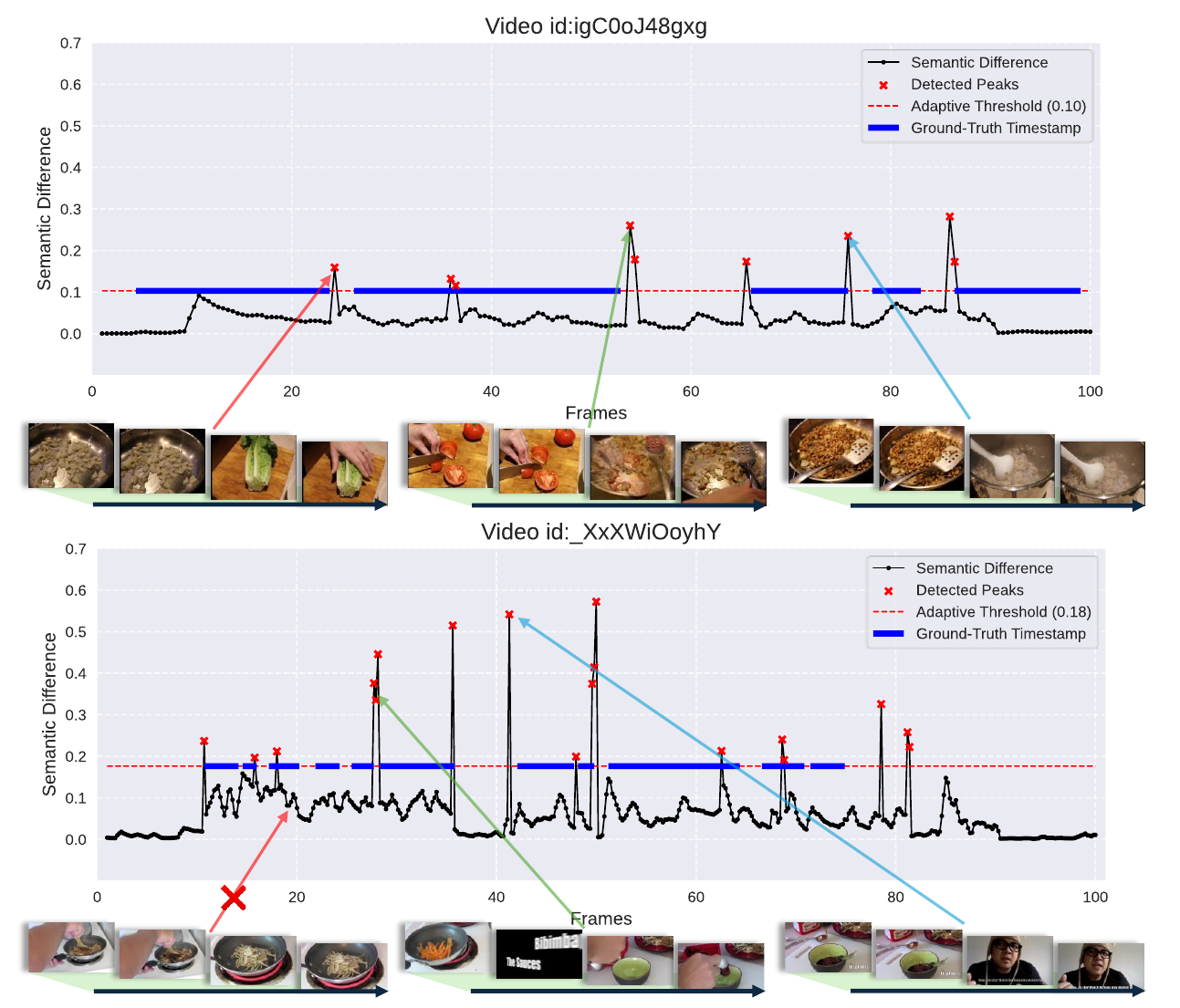}
\caption{Analysis for Semantic-Based Caption Retrieval.}
\vspace{-1mm}
\label{fig:semantic-analysis}
\end{figure*}

\Cref{fig:semantic-analysis} provides examples of how our semantic-based adaptive thresholding identifies meaningful segment boundaries based on semantic differences across frames. In the first example (Video id: \textit{igC0oJ48gxg}), our method successfully detects segment transitions that align well with visual changes in the cooking process, such as moving from chopping vegetables to frying. The detected peaks match the ground-truth timestamps, showing that the adaptive threshold (0.10) effectively captures event boundaries. In the second example (Video id: \textit{\_XxXWiOoYhY}), although some frames exhibit high semantic differences, our method correctly filters out false positives (red cross) that do not correspond to meaningful scene changes. The adaptive threshold (0.18) helps to focus on truly significant transitions, avoiding over-segmentation. These results demonstrate that our adaptive thresholding method dynamically adjusts to video content, effectively balancing sensitivity and precision in segment detection.

\section{More Qualitative Results.}
\label{supp:qualitative}
\begin{figure*}[h]
\centering\includegraphics[width=1\textwidth]{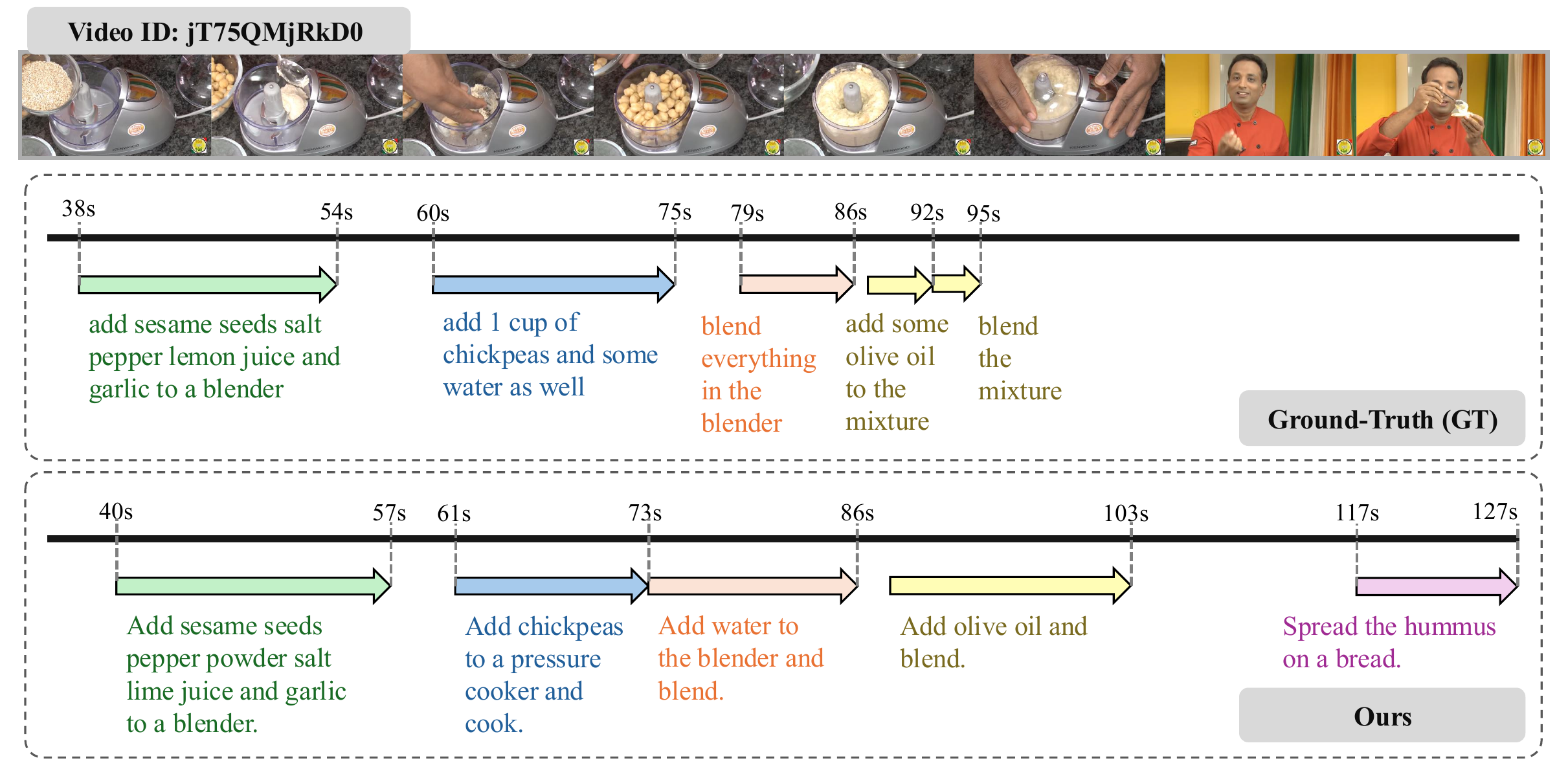}
\caption{More Qualitative Results for $\model$.}
\vspace{-1mm}
\label{fig:qualitative_sup}
\end{figure*}

We provide additional qualitative results in \Cref{fig:qualitative_sup} to further illustrate the effectiveness of our method.

% \section{Example Appendix}
% \label{sec:appendix}

% This is an appendix.

\end{document}